\documentclass[conference]{IEEEtran}
\IEEEoverridecommandlockouts
\usepackage{cite}
\usepackage{mathtools}
\usepackage{amsmath,amssymb,amsfonts,amsthm}
\usepackage{graphicx}
\usepackage{textcomp}
\usepackage{xcolor}
\usepackage{wrapfig}
\usepackage{caption}
\usepackage{subcaption}
\usepackage{url}

\definecolor{linkCol}{rgb}{0.2, 0.0, 0.1}
\definecolor{citeCol}{rgb}{0.1, 0.2, 0.3}
\definecolor{urlCol} {rgb}{0.2, 0, 0.5}
\usepackage[%
    raiselinks=true,  %
    colorlinks=true,  %
    plainpages=true,  %
    pdfpagelabels,    %
    citecolor=citeCol,urlcolor=urlCol,linkcolor=linkCol,
    unicode
]{hyperref}

\def\BibTeX{{\rm B\kern-.05em{\sc i\kern-.025em b}\kern-.08em
	T\kern-.1667em\lower.7ex\hbox{E}\kern-.125emX}}

\theoremstyle{definition}
\newtheorem{definition}{Definition}

\begin{document}

%\title{Train rescheduling based on feature space extracted from pairwise conflict resolution}

%\title{Optimizing train rescheduling based on feature space extracted from pairwise conflict resolution}

\title{Standardized feature extraction from pairwise conflicts applied to the train rescheduling problem}

%\title{Standardized feature extraction based on pairwise train scheduling conflicts}

%\title{A new approach to the train rescheduling problem based on feature space using pairwise conflict resolution}

\author{
    \IEEEauthorblockN{Anik\'o Kopacz}
	\IEEEauthorblockA{\textit{Faculty of Mathematics and Computer Science} \\
		\textit{Babe\c{s}-Bolyai University}\\
		Cluj-Napoca, Romania \\
		aniko.kopacz@ubbcluj.ro}
	\and
	\IEEEauthorblockN{\'Agnes Mester}
	\IEEEauthorblockA{\textit{Faculty of Mathematics and Computer Science} \\
		\textit{Babe\c{s}-Bolyai University}\\
		Cluj-Napoca, Romania \\
		\textit{\& Institute of Applied Informatics and Applied Mathematics} \\
		\textit{\'Obuda University}\\
		Budapest, Hungary \\
		agnes.mester@ubbcluj.ro}	
	\and
	\IEEEauthorblockN{S\'andor Kolumb\'an}
	\IEEEauthorblockA{\textit{Faculty of Mathematics and Computer Science} \\
		\textit{Babe\c{s}-Bolyai University}\\
		Cluj-Napoca, Romania \\
		sandor.kolumban@ubbcluj.ro}
	\and
	\IEEEauthorblockN{Lehel Csat\'o}
	\IEEEauthorblockA{\textit{Faculty of Mathematics and Computer Science} \\
		\textit{Babe\c{s}-Bolyai University}\\
		Cluj-Napoca, Romania \\
		lehel.csato@ubbcluj.ro}
}

\maketitle

\begin{abstract}
	We propose a train rescheduling algorithm which applies a standardized feature selection based on pairwise conflicts in order to serve as input for the reinforcement learning framework.
	We implement an analytical method which  identifies and optimally solves every conflict arising between two trains, then we design a corresponding observation space which features the most relevant information considering these conflicts. The data obtained this way then translates to actions in the context of the reinforcement learning framework. We test our preliminary model using the evaluation metrics of the Flatland Challenge. The empirical results indicate that the suggested feature space provides meaningful observations, from which a sensible scheduling policy can be learned. 
\end{abstract}


\section{Introduction}

Automated train rescheduling and traffic management are complex and challenging optimization problems, which are becoming increasingly pressing issues nowadays. Since several transportation and logistics companies are facing the need to increase their transportation capacity while maintaining effectiveness and reliability, lately there has been a growing interest in the development of train traffic optimization methods among researchers and companies such as the Swiss Federal Railways (SBB) or the Deutsche Bahn (DB).

Currently, train rescheduling is done by human dispatchers, which is evidently suboptimal and  unstable in terms of scalability and individual dependencies. As the problem is NP-complete (see e.g., \cite{Brucker}), the proposed approaches to its solution are diverse and many of them rely on heuristics. The most commonly used techniques include depth-first-search, branch and bound, genetic algorithms, tabu search and simulated annealing, see e.g., \cite{BranchAndBound, Genetic, DepthFirstSearch, SimulatedAnnealing} and references therein. For a comprehensive presentation of related work, see \cite{WangZhouLiu}.

Recently, substantial advances have been made in the field of reinforcement learning (RL) and, in particular, its application to automated traffic management. Several RL based approaches were implemented and tested for traffic signal control, see \cite{RLTraffic1, RLTraffic2, RLTraffic3} and references therein. In the context of train traffic optimization, \cite{Semrov_etAl} proposed a Q-learning algorithm on a single-track railway, \cite{RL2018} constructed a graph model of the railway infrastructure and applied a deep Q-network, while \cite{RoostetAl} used the A3C method (see \cite{Mnih_etAl}) in the first Flatland contest\footnote{See \url{https://www.aicrowd.com/challenges/flatland-challenge/submissions/26903}}. 

In 2019, SBB announced a train routing competition called the Flatland Challenge \cite{flatland:challenge2019}, which addresses the problem of efficient train traffic management on a complex railway infrastructure having dense traffic. This initiative has grown into a series of competitions\footnote{See the different editions of the Flatland Challenge at  \url{https://www.aicrowd.com/challenges/flatland-3}.}, from which the latest editions specifically promote the use of RL methods.

In the light of the above challenges, the purpose of this paper is to investigate a possible RL approach to the vehicle rescheduling problem formalized by the Flatland Challenge.
We propose a new feature extraction method by analyzing pairwise conflict situations, then we apply Q-learning (see \cite{SuttonBarto}) in the form of a neural Q-network, which plays the role of the decentralized controller of each train. We test our approach on a series of simulations, in accordance with the Flatland benchmark.

%The paper is organized as follows. 
The next section provides a brief description of the Flatland library and the adherent optimization problem. Section \ref{Proposed_Method} presents the studied approach to the Flatland Challenge, describing in detail the applied analytical method, the corresponding observation space and the reinforcement learning techniques in use. Section \ref{Experimental_Results} summarizes the validation results of the proposed method, by conducting a series of experiments in accordance with the Flatland Challenge evaluation metrics. Finally, section \ref{Conclusions} contains the analysis of the studied approach, as well as conclusions and comments regarding further directions of improvements.

\section{The Flatland Challenge} \label{Flatland_Challenge}

The Flatland library (see \cite{mohanty2020flatlandrl} and \cite{flatland:doc}) is an open-source framework implementing a multi-agent grid world environment, which was specifically designed to facilitate the development and testing of reinforcement learning algorithms for the rail transport system. This provides the framework of the Flatland Challenges, where the main goal is to create an adaptive controller which plans the trains’ routes and actions efficiently, and has the ability to adapt to possible delays and malfunctions.

\begin{figure}[htb]
	\begin{center}
		\captionsetup{justification=centering}
		\includegraphics[width=0.4\textwidth]{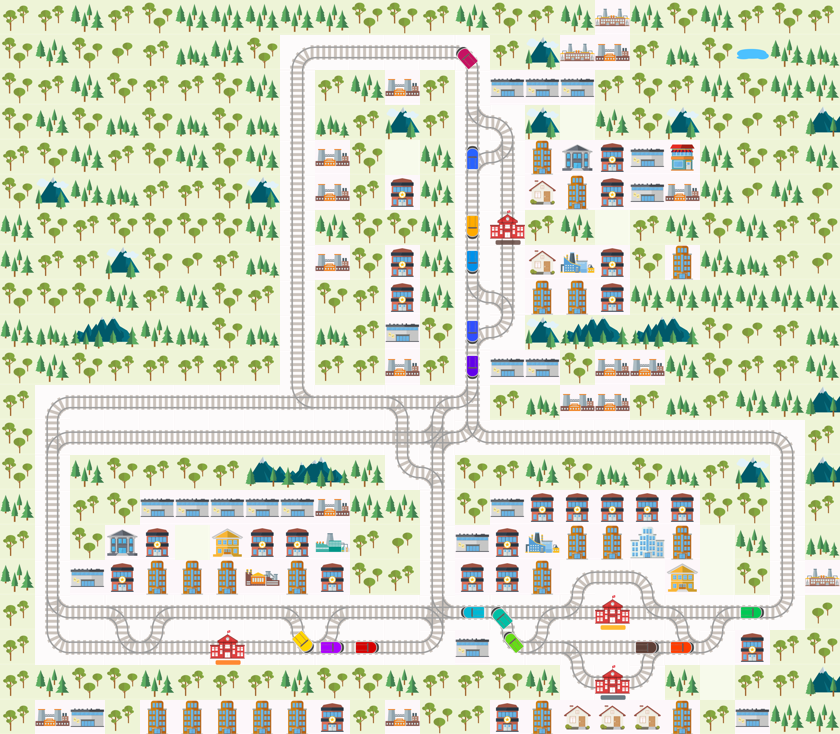}		
		\caption{Visualisation of a simple Flatland environment}
		\label{fig:env} 
	\end{center}
\end{figure}

Flatland is a discrete time simulator realizing a two-dimensional grid world, which consists of the railway infrastructure (i.e., the cells featuring the tracks, different switches and train stations), and the agents represented by the trains, see Figure \ref{fig:env}.

Each Flatland test environment consists of a randomized railway network and a set of trains, which are assigned a starting position, a constant speed value, a  predetermined moving direction and a target station.  
Given this configuration, the goal is to \textit{minimize the sum of travel time} of all  agents from their starting location to their respective destination.
In order to study this optimization problem, Flatland offers a public benchmark for experimenting with multi-agent RL algorithms, where the agents move by executing different actions, request environmental information under the form of observations, and get feedback by receiving rewards.

Generally, an agent can perform a limited set of actions: \textit{move forward}, \textit{turn left}, \textit{turn right}, \textit{stop}, or \textit{do nothing} (i.e., continue its previous action). 
However, in most cases only a few actions are valid, as the trains' possible movements are determined by the given railway infrastructure. This highly constrained action space proves to be one of the greatest challenges of the given optimization problem, as a single wrong decision can have irreversible and critical effects, manifested under the form of deadlocks. 

A \textit{deadlock} represents a situation where two or more trains cannot move from their current position because every train is blocked by at least one other train. Therefore, deadlocks should be avoided completely or reduced to a minimal amount, since the affected agents accumulate a significant penalty for not reaching their destination.  

The Flatland Challenge instantiates the so-called vehicle rescheduling problem (VRSP, \cite{Brucker}), which is an NP-complete optimization task. First of all, the growing number of trains, tracks and switches yields an exponential-growth of rerouting possibilities, which makes the global optimization of the full problem totally infeasible. Furthermore, the stochastic behaviour of the system caused by the disruption incidents adds an extra layer of complexity to the studied problem, which calls for constant monitoring and dynamic rescheduling. 
For more information regarding the action space, the reward signals, as well as the stochastic events manifested in the form of train malfunctions, see \cite{mohanty2020flatlandrl}, \cite{flatland:doc} and \cite{flatland:env}.

\section{Proposed Method} \label{Proposed_Method}

Our approach to the VRSP formalized by the Flatland Challenge relies on the identification, characterization and, if possible, resolution of every conflict arising between train pairs. Accordingly, we design the corresponding observation space featuring the most relevant information considering these detected conflicts. This data then provides an adequate state representation for the applied reinforcement learning framework.

\subsection{Problem decomposition} \label{section:A}

First, we decompose the given railway system into computationally smaller subproblems, which can be solved analytically with constant time complexity. Namely, we extract the family of  conflicts arising between \textit{two} trains; every individual conflict of this type can be solved optimally in $O(1)$ time, using an analytic formula obtained by solving the dynamic programming problem. 

The aforementioned conflicts create a covering of the whole railway environment, and the individual solutions of these problems can be used to construct feature vectors, which encode representative data regarding the state of the system. These observations then are used
as input for the neural network incorporated by the reinforcement learning model.

In order to reduce the magnitude of the computations, we remark that due to the typical infrastructure of railway networks, the vast majority of train positions do not require complex analysis in order to determine the agents' next actions. Namely, when an agent has only one possible next cell to move to, unless this cell is a switch, the default greedy action \textit{move forward} is favored and this will be taken, ensuring that the agent is not occupying resources any longer than it needs to.

Therefore, we can restrict ourselves to key situations where decision making is required, focusing on the actions to be taken right when an agent arrives to a junction or intersection.
In these cases, a reevaluation of the current state of the environment is necessary, in order to choose the most favorable action which  ensures the adaptive nature of the system.

Our proposed decomposition method relies on the following approach. Let us consider an agent which arrives to a critical position described above (i.e., its next cell is a switch). In this situation, we construct the shortest path from the train's current position to its predetermined destination for every possible maneuver which can be taken on the upcoming switch (i.e., \textit{turn left},  \textit{move forward}, or \textit{turn right}). This results in maximum $3$ distinct routes for one particular agent, all of which are determined by the $A^*$ algorithm implemented in the Flatland library. We repeat the process for every other train, constructing their corresponding shortest paths by using the first switch which they are going to encounter by successively taking the action \textit{move forward}. Then, we pair the possible paths of the agent in question with every other agent's set of routes, and we identify which path pairs intersect with each other. We mark the respective intersecting agent routes as possible conflicts and give an analytic solution to them, based on the trains' current positions, velocities, length of the critical section and the distances from the nearest switches.

% --------------------------------------

\subsection{Analytical solution of pairwise problems}

By analyzing the possible train pair dispositions, we notice that all the potential conflicts arising between two trains can be represented by a finite number of specific scenarios. Therefore, we propose analytical solutions to each of these particular situations in order to determine the optimal actions for the involved agents.

We stipulate that every arbitrary conflict between two given trains can be modeled by a simplified scenario called the \textit{H-shaped Flatland} problem. Let $T_1$ and $T_2$ be two  trains paired with their precalculated routes $\mathcal{P}(T_1)$ and $\mathcal{P}(T_2)$, which are obtained by the application of the $A^*$ algorithm (see the previous section). In order to develop a unified analytical approach to the potential train pair conflicts, we introduce the following notions:  

\begin{definition}
	We say that $T_1$ and $T_2$ are \textit{in conflict} if there exists a common section of the paths $\mathcal{P}(T_1)$ and $\mathcal{P}(T_2)$, i.e., the set of cells $\mathcal{M} \coloneqq \mathcal{P}(T_1) \cap \mathcal{P}(T_2)$ is not empty.
\end{definition}

\begin{definition}
	Let $T_1$ and $T_2$ be two trains in conflict.
	The \textit{critical section} $\mathcal{C}$ associated with the ordered pair $(T_1,T_2)$ is the shortest railway section delimited by switches/ train stations on both ends, which contains the first connected segment of the intersection $\mathcal{M} = \mathcal{P}(T_1) \cap \mathcal{P}(T_2)$ closest to $T_1$.
	We denote the endpoints of $\mathcal{C}$ by $S_1$ and $S_2$. 
\end{definition}

Note that the intersection $\mathcal{M}$ and the critical section $\mathcal{C}$ do not necessarily coincide. The critical segment $\mathcal{C}$ can be determined based on $\mathcal{M}$, first by isolating the connected subroute of the set $\mathcal{M}$ closest to $T_1$, then expanding this section in order to obtain a segment delimited by switches/train stations on both ends. 
If the endpoints of the connected subpath are already switches/ stations, there is no need for expansion. We denote the length of the critical section $\mathcal{C}$ by a natural number $l \geq 1$, indicating the number of cells present in the corresponding segment. Note that in the particular case $l=1$, $\mathcal{C}$ is composed of a single switch, i.e., $S_1$ and $S_2$ coincide.

\begin{figure}[htb]
	\captionsetup{justification=centering}
	\captionsetup[subfigure]{justification=centering}
	\begin{subfigure}{0.241\textwidth}
		\includegraphics[width=\textwidth]{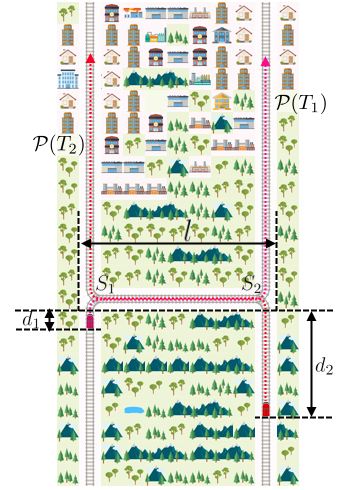}		
		\caption{Opposite directional conflict}
		\label{fig:Hshape_opposite}
	\end{subfigure}
	\hfill
	\begin{subfigure}{0.241\textwidth}
		\includegraphics[width=\textwidth]{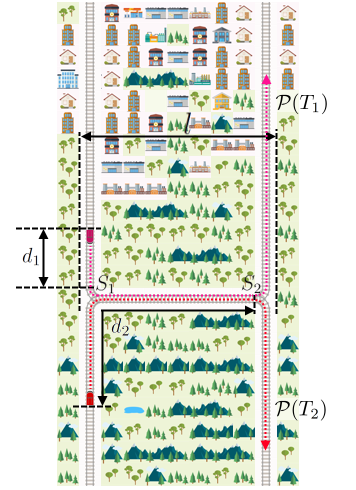}			
		\caption{Same directional conflict}
		\label{fig:Hshape_same}
	\end{subfigure}
	
	\caption{The H-shaped Flatland model}
	\label{fig:Hshape}
\end{figure}

This expansion technique leads to the so-called \textit{H-shaped Flatland model}, whose topology is illustrated in Figure \ref{fig:Hshape}. In this environment we consider two trains $T_1$ and $T_2$, each having an arbitrary starting position, moving direction, and a fixed speed -- $v_1$ and $v_2$. These speed profiles are implemented as constant fractions in the $(0,1]$ interval, see \cite{flatland:doc}. The critical section $\mathcal C$ is given by the single-track railway in the middle, being delimited by the switches $S_1$ and $S_2$.

The H-shaped problems can be divided into two families, depending on whether the conflicting trains have \textit{opposite} or \textit{similar  direction} of movement in the critical section (see Figures \ref{fig:Hshape_opposite} and \ref{fig:Hshape_same}). Note that there exists a major difference between the two classes, since  in the case of opposite directions, deadlock may form in the critical section when not taking the appropriate actions, which is not the case in the same directional problem.  

Let $d_1$ be the \textit{signed} distance between train $T_1$ and the switch $S_1$, i.e., $d_1$ represents the number of cells between $T_1$ and $S_1$, and $d_1$ is nonnegative if $S_1$ belongs to the planned path $\mathcal P(T_1)$, otherwise  $d_1<0$. In the particular case $d_1=0$ the train $T_1$ is located on the cell of the switch $S_1$. The number $d_2$ is defined similarly. The sign of the distances is an important factor in the deduction of the trains' optimal actions, e.g., in the opposite directional case, the condition $d_2\leq0$ indicates that $T_2$ is already occupying the critical section, therefore $T_1$ must stop when arriving to the switch $S_1$ in order to prevent the appearance of a deadlock.        

Considering these H-shaped scenarios and setting aside the possibility of train disruptions, the optimal policy of the two agents can be determined explicitly and with constant time complexity based on the variables $d_1$, $d_2$, $l$, $v_1$ and $v_2$.
Note that, in accordance with the  evaluation metrics of the Flatland Challenge, we define the optimal policy as a policy that maximizes the reward signal, i.e., minimizes the cumulative travel time of all agents from their current position to their target destination.

Now let us derive the optimal policy of $T_1$ in case of the opposite directional problem, see Figure \ref{fig:Hshape_opposite}. In concordance with the greedy principle detailed in the previous section, the optimal action of the agent is to \textit{move forward}, except when $T_1$ arrives to a critical position requiring decision, i.e., the next cell of its path is the switch $S_1$. In the current Flatland version \cite{flatland:doc}, this location is characterised by the condition $d_1 = 1$. In this case, if $d_2 \geq 1$ then the optimal action of $T_1$ is given by $\text{argmin}(t_1^o, t_2^o)$, where 
\begin{equation} \label{eq:t1}
	t_1^o = \frac{d_1+l}{v_1} + \max\left(\frac{d_1+l}{v_1}, \frac{d_2-1}{v_2}\right) + \frac{l+1}{v_2}
\end{equation}
and			
\begin{equation} \label{eq:t2}
	t_2^o = 2 \frac{d_2+l}{v_2} + \frac{d_1+l}{v_1}.
\end{equation}

However, when $d_1 = 1$ and $d_2 < 1$, then $T_1$ is required to \textit{stop} in order to avoid a deadlock.  
Note that this policy generalizes well for the signed distances: when $d_1 \leq 0$, the optimal action of $T_1$ is still to \textit{move forward} on its path $\mathcal{P}(T_1)$ in order to liberate the critical section as soon as possible; if $d_2 \leq 0$ as well, then this disposition represents an inevitable deadlock situation.

The same directional conflict (see Figure \ref{fig:Hshape_same}) can be solved in a similar manner, depending on the parameters  $d_1$, $d_2$, $l$, $v_1$ and $v_2$. Again, the optimal action for $T_1$ is to \textit{move forward}, unless $T_1$ arrives to a location requiring decision, which is given by the condition $d_1 = 1$. In this case, first we determine the conflicting position $P$ of $T_1$ and $T_2$, i.e., the cell belonging to the critical section $\mathcal{C}$ where $T_1$ and $T_2$ would meet at the same time, given their current trajectories. If $P \notin \mathcal{C}$, then $T_1$ should \textit{move forward}, otherwise the optimal action can be determined by $\text{argmin}(t_1^s, t_2^s)$, where
\begin{equation} \label{eq:t1_same}
	t_1^s = 2\frac{d_1+l}{v_1}
\end{equation}
and			
\begin{equation} \label{eq:t2_same}
	t_2^s = \frac{2 d_2-l}{v_2} + \frac{d_1+l}{v_1}.
\end{equation} 

The particular case $l=1$ can be solved analogously in both the opposite and same directional scenario. 

The described analytical solutions give a characterization of the conflicting train pair system, and the estimated times associated to the corresponding actions (see relations \eqref{eq:t1}--\eqref{eq:t2_same}) provide adequate observations of the respective agents, which will be used as feature vectors in the reinforcement learning context.

%TODO - kiegesziteni a rajzot: rairni S_1, S_2 switcheket

\subsection{Design of the observation space}
\label{sec:DesignOfObservationSpace}

We propose to construct the agents' observations  by successively applying the H-shaped Flatland model. Namely, for every agent whose next cell is a switch, we construct a $4$-dimensional vector  encoding numerical information about the desirability of its $4$ actions which can be executed upon arriving to the switch (i.e., \textit{move forward}, \textit{turn left}, \textit{turn right} or \textit{stop}). We discard the \textit{do nothing} action in order to reduce the dimensionality of the problem. 

The observation vectors are determined in the following way: when an agent $T_i$ is facing a switch, for every possible action which can be made on the upcoming junction, we construct its shortest path $\mathcal{P}(T_i)^k$ to its target destination ($k \in {1,2,3}$ representing the potential actions), then we pair these routes with the maximum $3$ shortest routes $\mathcal{P}(T_j)^m$ of every other train $T_j$ determined in the way described in section \ref{section:A}. Then, for every train pair $(T_i, T_j)$ and their fixed paths $\mathcal{P}(T_i)^k$ and $\mathcal{P}(T_j)^m$, we determine whether the two trains are in conflict, and if so, we fit the H-shaped model to the given setting, as described in the previous section. 
Finally, we label every component of the feature vector $v^i \in \mathbb{R}^4$ with a numerical value which represents the normalized estimated arrival time of $T_i$ corresponding to that specific action. 

In particular, for $k \in \{1,2,3\}$, i.e., for the actions \textit{move forward}, \textit{turn left} and \textit{turn right}, we have 
$$v_k^i = \frac{1}{|\mathcal{P}_{i,k}|} \sum_{\mathcal{P}(T_j)^m \in \mathcal{P}_{i,k}} t_{(\mathcal{P}(T_i)^k, \mathcal{P}(T_j)^m)},$$
where the set $\mathcal{P}_{i,k}$ denotes the trains and their corresponding paths which are in conflict with the route $\mathcal{P}(T_i)^k$, such that the conflict resolution results in the $k^{\text{th}}$ action for agent $T_i$. The number $t_{(\mathcal{P}(T_i)^k, \mathcal{P}(T_j)^m)}$ represents the estimated arrival time of $T_i$ to its target destination, considering that $T_i$ takes the $k^{\text{th}}$ action on the upcoming switch and follows the route $\mathcal{P}(T_i)^k$. The respective time can be obtained by using the formula given by the H-shaped Flatland model (see relations \eqref{eq:t1} and \eqref{eq:t1_same}).

For the action \textit{stop}, we set
$$v_4^i = \min_{k=\overline{1,3}} \max _{\mathcal{P}(T_j)^m \in \mathcal{P}^s_{i,k}} t_{(\mathcal{P}(T_i)^k, \mathcal{P}(T_j)^m)},$$
where the set $\mathcal{P}^s_{i,k}$ represents the conflicts of $T_i$ for which the application of the H-shaped technique dictates that the optimal action of $T_i$ is to \textit{stop} upon arriving to the switch. The times $t_{(\mathcal{P}(T_i)^k, \mathcal{P}(T_j)^m)}$
are calculated by taking into account the waiting periods of $T_i$, which are also given by the H-shaped model (see relations \eqref{eq:t2} and \eqref{eq:t2_same}).

\vspace{0.5cm}

\subsection{The neural network}

The observation vectors built from the characterization of the pairwise conflicts constitute the input to a neural network, which provides the decentralized controller for the scheduling problem. 
We adopt Q-learning, an off-policy model-free reinforcement learning technique, for approximating the common action-value function of the agents. We test our method by applying deep Q-networks (DQN, \cite{mnih_dqn}) and a different network architecture referred to as linear Q-network (LQN), as well.

In case of the DQN approach, we utilize the same network architecture for both the value- and advantage-network: the networks consist of two hidden linear layers with the size of $64$ and ReLU activation. 

On the other hand, the LQN approach applies a single perceptron layer in order to approximate the value function and the advantage function, respectively.

\section{Experimental results}
\label{Experimental_Results}

We test a preliminary version of our approach in the Flatland framework, over a series of $28 \times 16$  test environments, where the number of agents is set to $6$, the number of train stations is $4$, and train malfunctions are not permitted. All agents operate with the same speed, they traverse $1$ cell each time step.

We evaluate the training results of the applied LQN and DQN approaches, then we compare these methods with the greedy baseline policy, which follows every agent's shortest path determined by the $A^*$ algorithm, disregarding other trains.  The evaluation metrics are given by the accumulated reward signals obtained during the training process, the length of the generated episodes, the number of agents and the percentage of agents that successfully arrived to their target station; these metrics are all in accordance with the Flatland Challenge benchmark, see \cite{flatland:metrics}.

\begin{figure}[htb]
	\begin{center}
		\captionsetup{justification=centering}
		\includegraphics[width=0.38\textwidth]{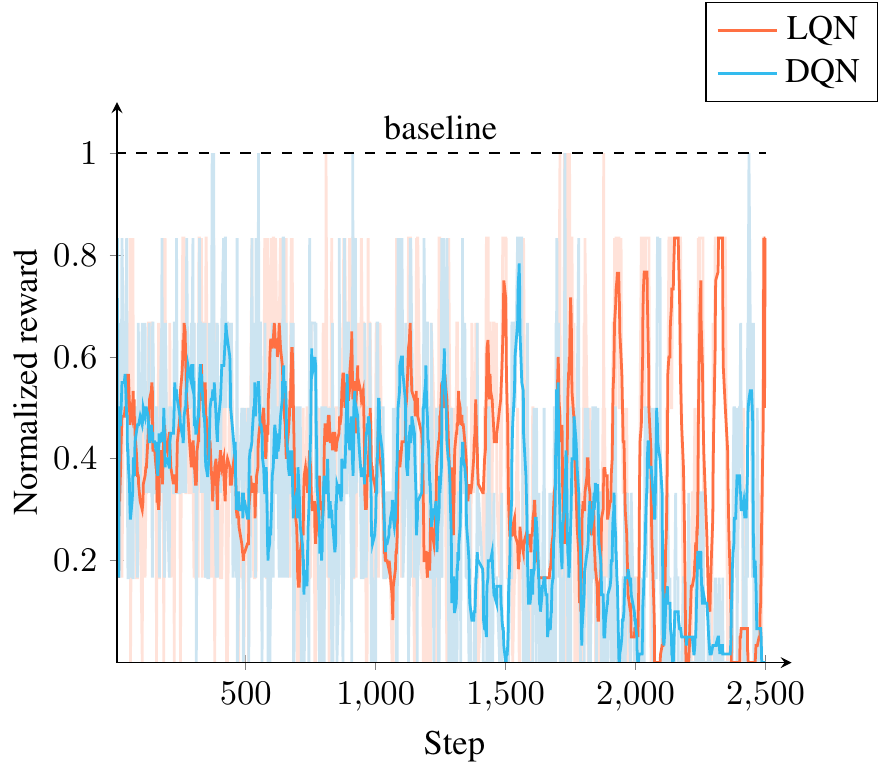}		
		\caption{Evaluation of the training results of our proposed method based on the normalized reward values, in comparison with the baseline algorithm.}
		\label{fig:results}    
	\end{center}
\end{figure}

Figure \ref{fig:results} illustrates the performance of the three algorithms based on the normalized accumulated reward values of the episodes, which are determined as 
$$\overline{r} = 1 + \frac{r}{T \cdot N
},$$ where $r$ is the reward signal accumulated during a given episode, while $T$ and $N$ are the total number of time steps and the number of agents of that particular episode. 
We applied a moving average to the original normalized reward values in order to analyze the overall tendencies of the methods. % keep it or leave it sentence

%The preliminary model approaches the baseline policy in the case of small-scale problems, identifying and solving the majority of critical situations successfully. However, it seems to fail at adapting efficiently to environments with high agent density. 
%
The preliminary model approaches the baseline policy for some environmental scenarios in case of small-scale problems, identifying and solving the majority of critical situations successfully. However, it seems that even for environments with low agent density the system is exposed to the emergence of deadlocks. 

%
%The combination of reinforcement learning methods with the  proposed feature extraction is capable of avoiding unfavorable decisions and ensuring that the agents reach their destination, even if it does not always determines the optimal solution. 

 %less sensitive to the complexity of the system. 

\section{Conclusions}
\label{Conclusions}

\subsection{Evaluation of the experiment}
\label{sec:EvaluationOfTheExperiment}

The goal of the present work was to introduce a standardized framework for constructing feature vectors to be used in machine learning algorithms in relation to train rescheduling problems. The presented method relies on the optimal resolutions of pairwise conflicts. There is an unbounded number of ways in which the information extracted from pairwise conflicts can be combined into a feature vector. We have given one such example in Section \ref{sec:DesignOfObservationSpace} and attempted learning a scheduling policy using this feature space, the results of which are presented in Section~\ref{Experimental_Results}. The advantage of the proposed feature selection algorithm is that the output is not dependent on the parameters of the environment, thus ensuring the possibility to train and evaluate on agent configurations with different magnitudes.

The experimental results illustrate that a meaningful scheduling algorithm can be learned even based on such a simple feature space. This is supported by the fact that in many cases the obtained normalized reward reaches values above 0.7 (see Fig.~\ref{fig:results}).
However, it needs to be underlined that these results are by far not sufficient from any practical point of view.
 
%The ordering of test cases in Fig.~\ref{fig:results} can be thought of as an order by complexity. While on the first half of the test cases the obtained LQN and DQN models perform similarly,  more and more complex Flatlands result in the DQN model getting visibly less reward compared to the LQN. 

\subsection{Future work}
\label{sec:FutureWork}

During the evaluation of different feature spaces constructed by applying the introduced framework, we have already seen that naive choices perform really badly, not offering the possibility to learn a sensible scheduling model. Even the most trivial problems can trip neural networks trained by using these poor feature spaces. Part of our future efforts will go into finding ways to translate information extracted from pairwise conflict resolutions into features. We have already learned quite a lot on this front by analyzing the failures of the naive feature spaces, and this knowledge needs to be expanded much further in order to achieve and surpass the reward collection capabilities of the baseline algorithm.

Before applying such an algorithm to real life situations, it is of utmost importance to verify their performance on historic scenarios. Through a collaboration with a partner of the SBB, we will try to collect and construct a training dataset containing the description of H-shaped conflicts from the history of the Swiss railways. Learning and testing based on those scenarios would be very valuable in gaining further insights.

Last but not least, the presented experiment already showed significant differences between the tested neural network architectures. Another focus area of our future investigations will be to compare different architectures.

\section*{Acknowledgment}

The work of Anikó Kopacz and Lehel Csató was supported by the Domus Program of the Hungarian Academy of Sciences, grant nr. 1946/22/2021/HTMT.

{ %\renewcommand{\baselinestretch}{0.8}
	%	\normalsize 
	%	\setlength{\itemsep}{-2.4mm}
	\bibliographystyle{abbrv}
	\bibliography{bibliography}
}

\end{document}